\documentclass[10pt,twocolumn,letterpaper]{article}

\usepackage[pagenumbers]{cvpr}

\usepackage[utf8]{inputenc}
\PassOptionsToPackage{square,numbers}{natbib}

\usepackage[utf8]{inputenc} 
\usepackage[T1]{fontenc}    
\usepackage[skip=10pt]{caption}
\usepackage{url}            
\usepackage{booktabs}       
\usepackage{amsfonts}       
\usepackage{nicefrac}       
\usepackage{microtype}      
\usepackage{algorithm}
\usepackage[noend]{algpseudocode}
\usepackage{graphicx}
\usepackage{xcolor}
\usepackage{amsmath}
\usepackage{MnSymbol}
\usepackage{xcolor}
\definecolor{darkgreen}{RGB}{0,128,0}

\usepackage[pagebackref,breaklinks,colorlinks]{hyperref}

\usepackage[capitalize]{cleveref}
\crefname{section}{Sec.}{Secs.}
\Crefname{section}{Section}{Sections}
\Crefname{table}{Table}{Tables}
\crefname{table}{Tab.}{Tabs.}

\title{
A Picture is Worth a Thousand Words: Principled Recaptioning Improves Image Generation}
\author{Eyal Segalis\\
Google Research\\
{\tt\small eyalis@google.com}
\and
Dani Valevski\\
Google Research\\
{\tt\small daniv@google.com}
\and
Danny Lumen\\
Google Research\\
{\tt\small dwasserman@google.com}
\and
Yossi Matias\\
Google Research\\
{\tt\small yossi@google.com}
\and
Yaniv Leviathan\\
Google Research\\
{\tt\small leviathan@google.com}
}\date{September 2023}

\begin{document}
\twocolumn[{
\maketitle
\begin{center}
    \centering
    \captionsetup{type=figure}
    \includegraphics[width=0.95\linewidth]{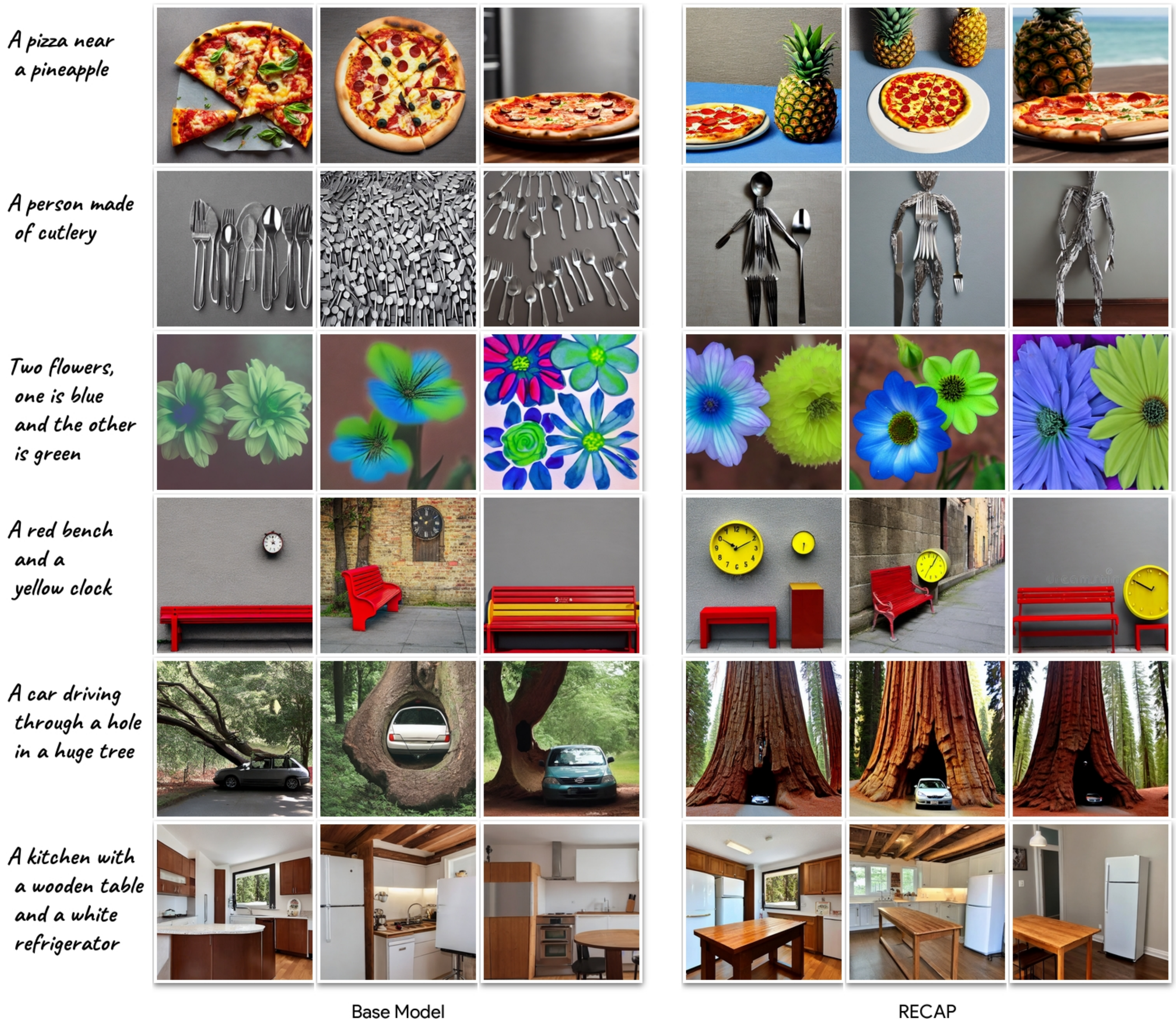}
    \captionof{figure}{Examples of non-cherrypicked generations from the base Stable Diffusion model (left) and our model trained on a recaptioned dataset (right), on the same set of random seeds.}
    \label{fig:top_example}
\end{center}
}]

\section{Abstract}
Text-to-image diffusion models achieved a remarkable leap in capabilities over the last few years, enabling high-quality and diverse synthesis of images from a textual prompt. However, even the most advanced models often struggle to precisely follow all of the directions in their prompts. The vast majority of these models are trained on datasets consisting of (image, caption) pairs where the images often come from the web, and the captions are their HTML alternate text. A notable example is the LAION dataset, used by Stable Diffusion and other models. In this work we observe that these captions are often of low quality, and argue that this significantly affects the model's capability to understand nuanced semantics in the textual prompts. We show that by relabeling the corpus with a specialized automatic captioning model and training a text-to-image model on the recaptioned dataset, the model benefits substantially across the board. First, in overall image quality: e.g. FID 14.84  vs. the baseline of 17.87, and 64.3\% improvement in faithful image generation according to human evaluation. Second, in semantic alignment, e.g. semantic object accuracy 84.34 vs. 78.90, counting alignment errors 1.32 vs. 1.44 and positional alignment 62.42 vs. 57.60. We analyze various ways to relabel the corpus and provide evidence that this technique, which we call RECAP, both reduces the train-inference discrepancy and provides the model with more information per example, increasing sample efficiency and allowing the model to better understand the relations between captions and images.

\section{Introduction}
In recent years, text-to-image (T2I) generation models such as Imagen \cite{Imagen}, Muse \cite{muse}, Dall-E \cite{DALLE1}, Dall-E 2 \cite{DallE2}, Parti \cite{Parti}, and Stable Diffusion \cite{rombach2022highresolution} have undergone significant advancements. This progress has enabled the generation of remarkably high-quality and diverse images by conditioning on textual inputs. However, while revolutionary, even modern state-of-the-art text-to-image models may fail to generate images that fully convey the semantics and nuances from the given textual prompts. Failure modes include: missing one or more subjects from the input prompt \cite{Parti, chefer2023attendandexcite}; incorrect binding of entities and modifiers \cite{Parti, rassin2023linguistic, chefer2023attendandexcite}; and incorrect placement and spatial composition of entities \cite{Parti, wu2023harnessing, phung2023grounded}.

In this work we first observe that open-web datasets used to train open text-to-image models suffer from significant issues. For example, the captions in the LAION \cite{schuhmann2022laion5b} dataset, used to train Stable Diffusion, come from alt HTML tags (Alttext). According to W3C's web content accessibility guidelines\footnote{\url{https://www.w3.org/TR/2016/NOTE-WCAG20-TECHS-20161007/H37}}, the alt attribute is used to convey the meaning and intent of the image, and not necessarily being a literal description of the image itself. Indeed, we observe, that often the Alttext describes only a narrow aspect of the image, neglecting significant visual details. For example, an image of a person can have as Alttext the name of the person and the name of the photographer, but not a description of their appearance, their clothes, their position, or the background. Also, sometimes Alttext tags contain inaccuracies, mistakes and out of context information. See \cref{fig:pali_outputs} for examples.

We further observe that while trained mainly on similar datasets of open (image, caption) pairs, recent automatic captioning systems, such as PaLI \cite{chen2023pali}, produce highly accurate captions. See examples in \cref{fig:pali_inputs}. This may be due to the fact that the inverse problem of image-to-text (I2T) is easier, or to the fact that these captioning models are larger, are trained longer than the T2I models, or leverage large pre-trained language components.

With these observations, we suggest a new method for horizontally improving T2I models by training them on improved captions, auto-generated by a custom I2T model. We call our method RECAP, and show that applying it to Stable Diffusion  results in a model that is better than the baseline across the board, with a battery of standard metrics substantially improving, e.g. FID  $17.87{\rightarrow}14.84$, as well as in human evaluation of successful image generation $29.25\%\rightarrow48.06\%$ (see Section \ref{results}).

In \cref{method} we provide the details of our method, RECAP. \cref{results} discusses our results and shows the horizontal improvements both in image quality as well as in semantic fidelity to the prompts.
In \cref{analysis} we analyze the issues with the original captions, demonstrate that the improvements are indeed due to the new captions, and that they arise as a result of both minimizing the train-test skew as well as increasing sample efficiency.

To summarize, our main contributions are:
\begin{itemize}
\item A new method we call RECAP, that leverages automatic captioning to improve the quality of a text-to-image model in a substantial way horizontally, both in fidelity and semantics, measured on a set of 7 standard metrics as well as with human evaluations.
\item An analysis showing how Alttext captions used by current training methods suffer from train-inference skew and lack in semantic details, resulting in text-to-image models that often fail in fidelity and semantic alignment, and how different captions mitigate both issues.
\end{itemize}

\begin{figure*}[htbp]
    \centering
    \includegraphics[width=1.\linewidth]{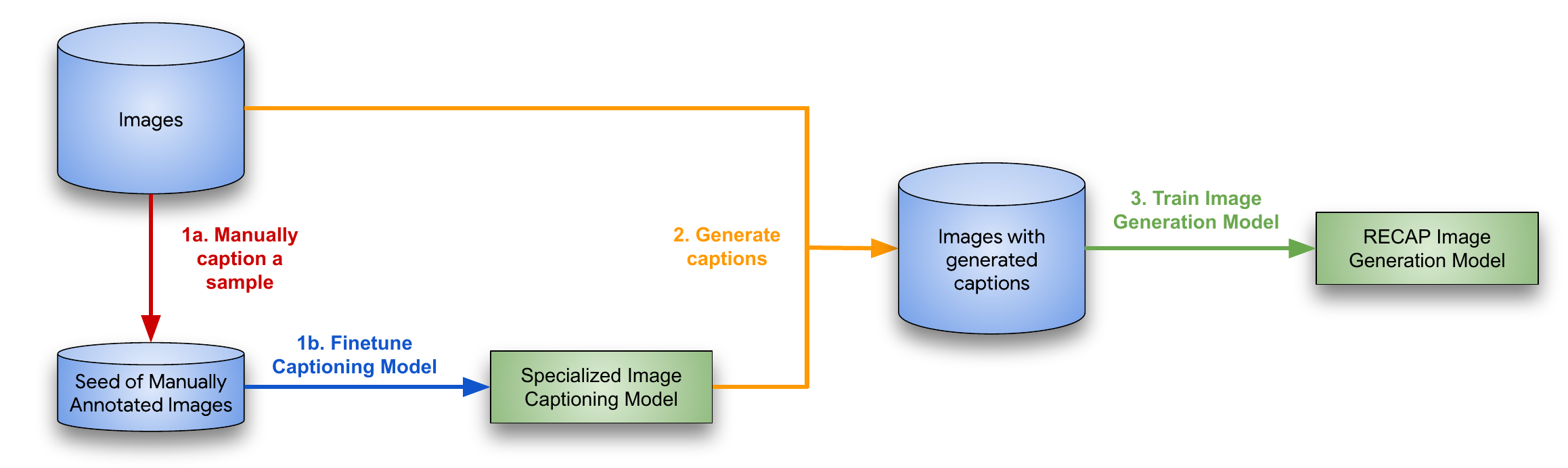}
    \captionof{figure}{Schematic diagram of our method RECAP.  In steps (1a) and (1b) we fine-tune an image-to-text captioning model on a small set of detailed human captions. In step (2) we use this fine-tuned model to recaption the images in the training dataset of a text-to-image model, and with this dataset, in step (3) we train an image generation model with the recaptioned dataset.}
    \label{fig:diagram}
\end{figure*}

\section{Related Work}
\textbf{Text-to-Image Models.} Deep generative models for image generation from text have shown notable progress in recent years, transitioning from using GAN-based methods \cite{GAN} to using autoregressive transformers \cite{DALLE1,Parti,muse} and diffusion models \cite{DallE2,StableDiffusion,Imagen}. An important area of enhancement is in improving a model's capability to align with the input text effectively. Methods condition the image on the output of a pre-trained text embedder, usually CLIP \cite{CLIP}. Imagen \cite{Imagen} shows that using a strong T5 \cite{T5} encoder significantly improves text-image alignment. Parti \cite{Parti} uses a SimVLM model \cite{simvlm} to annotate some of the training set images.

\textbf{Image captioning.} Image captioning is a fundamental problem in computer graphics. Recent models \cite{li2023blip2,alayrac2022flamingo,yu2022coca,chen2023pali,llava} have made significant progress on this task, thanks to high scale training data and utilizing pre-trained image and text models, allowing them to address a variety of multi-modal tasks. In this work we fine-tune PaLI \cite{chen2023pali} to recaption the training dataset of a text-to-image model.

\textbf{Synthetic multimodal datasets.} Some concurrent works augment multimodal dataset via automatic means to improve the capabilities of image captioning and embedding models. Nguyen et al. \cite{nguyen2023improving} show that training CLIP on data with generated captions improves its performance. Li et al. \cite{li2023data} show how to improve the BLIP model by iteratively removing captions with high loss, replacing them with captions from the previous epoch or generating a new image for them using Stable Diffusion. Ma et al. \cite{ma2023textonly} generate a synthetic text-to-image model using Stable Diffusion and then uses it to train a captioning algorithm.

\textbf{Improving text-alignment of diffusion models.} An important line of work attempts to rectify text-to-image alignment issues in diffusion models. This is usually done by lexical analysis of the input prompt modifying the attention maps throughout sampling. Chefer et al. \cite{chefer2023attendandexcite} try to prevent the diffusion models from ignoring certain tokens by guiding the sample process to reweigh the token attention maps. Rassin et al. \cite{rassin2023linguistic} analyze the prompt to identify modifiers, and then uses a sample-time guidance to bind their attention maps to those of their entities. Wu et al. \cite{wu2023harnessing} isolates phrases that describe individual entities and attends to them. Phung et al. \cite{phung2023grounded} allows placing objects in certain region by boosting their attention in those regions.

Concurrently with our work, Dall-E 3 \cite{dalle3} proposes to use an automatic captioning system to regenerate the captions used to train a T2I model. Our work uses an open model (Stable Diffusion) and we provide more details and focus more on analysis and evaluation, but otherwise the main ideas are very similar.

\section{Method} \label{method}

Our method, RECAP consists of 3 steps: (1) fine-tune an automatic captioning system to produce desirable labels; (2) relabel the images from our text-to-image training set with this automatic recaptioning system; and (3) train the text-to-image model on the dataset consisting of the images and new captions. \cref{fig:diagram} visualizes the overall method.

\subsection{Training Dataset}
We selected a subset of 10M photos from \emph{LAION-2B-en improved Aesthetics} dataset. We followed the data filters used to train Stable Diffusion versions 1.2-1.4\footnote{\url{https://huggingface.co/CompVis/stable-diffusion}}, as follows: $aesthetics\ score \ge 5.0,\ pwatermark < 0.5,\ nsfw$ is $UNLIKELY$ and both $height$ \& $width \ge 512$. We note that this filtering operation might amplify biases in the dataset \cite{birhane2021multimodal}.
We further excluded an arbitrary subset of 10K photos from the training set to be used as an internal validation set.

\subsection{Captioning Model} \label{methods_pali_finetune}
We used a pre-trained I2T captioning model (PaLI \cite{chen2023pali}). As the model outputs are relatively terse and lack in detail, we first collected a small set of 100 manual captions from human raters and fine-tuned the captioning model on that set.

\begin{figure*}[htbp]
    \centering
    \includegraphics[width=1.0\linewidth]{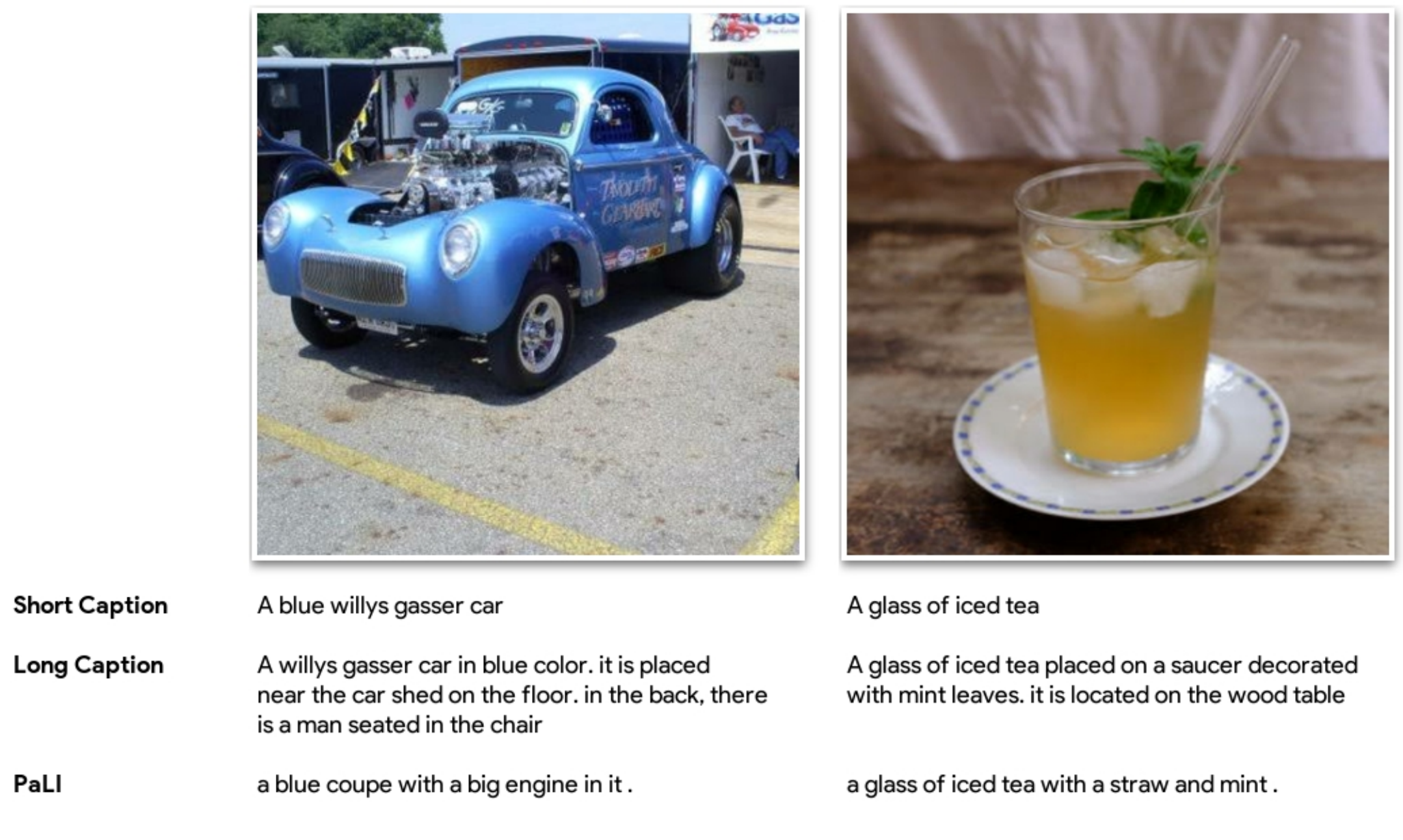}
    \captionof{figure}{Examples of captions given by human raters, and the automatically generated caption from the non fine-tuned PaLI model. Photos taken from LAION.}
    \label{fig:pali_inputs}
\end{figure*}

In order to experiment with the effect of different captioning distributions, the raters were asked to provide two types of captions. First, a detailed caption for each image, with these instructions: ``Describe what you see in each image using 1-2 detailed sentences''. Note that the instructions limited the length to only 1-2 detailed sentences at most, due to CLIP's (the downstream text encoder) context size of only 77 tokens, which longer captions exceeded. Second, we collected a short and less detailed caption for each image with this instruction: ``Describe what you see in each image using a single short sentence''.
We did not iterate on the quality of these manual captions. \cref{fig:pali_inputs} provides example captions given by the human raters, as well as those produced by the non fine-tuned PaLI model.

With this small dataset, we fine-tuned PaLI for 300 steps, using a learning rate of 4e-5, dropout rate of 0.1 and a batch size of 64, mixing 50\% short captions and 50\% long captions for multiple copies of the 100 images, using a different fixed conditioning prefix for the short vs. long captions. When generating captions from the fine-tuned model, we use the terms \emph{RECAP Short} and \emph{RECAP Long} to refer to generations conditioned on the short and long prefixes respectively. Example outputs, as well as a comparison to the original Alttext captions can be found in \cref{fig:pali_outputs}.
See \cref{appendix_pali_outputs} for additional dozen random examples of the generated RECAP captions.

Overall, the captions produced by this bespoke model improve both of the issues above - their  distribution better matches inference time prompts and they contain much more detail to improve sample efficiency (see \cref{ablation_caption_distribution_captions}).

\begin{figure*}[htbp]
    \centering
    \includegraphics[width=1.0\linewidth]{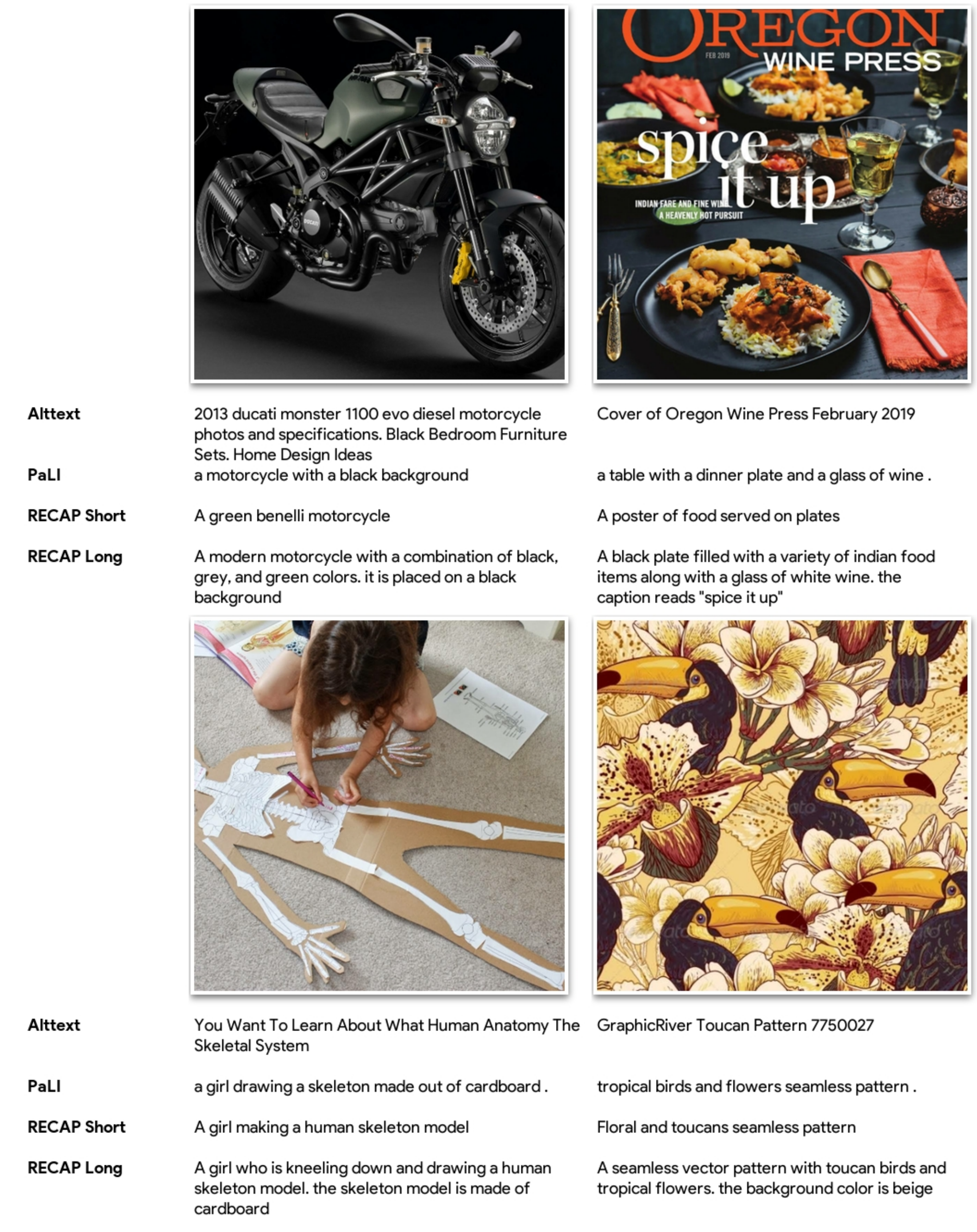}
    \captionof{figure}{Examples of captions generated by the RECAP model conditioned on the short or long prefixes, the original PaLI model, and the original Alttext captions. Photos taken from LAION.}
    \label{fig:pali_outputs}
\end{figure*}

\subsection{Image Generation Model}
Next, we fine-tuned Stable Diffusion v1.4 for an additional 250k\footnote{When further fine-tuning the model for 1M steps, we observe better visual results along with diminishing returns in the auto computed metrics. All results in the paper are for 250K steps except Fig. 1 and the human evaluation, where we used a model fine-tuned for 1M steps.} steps with a learning rate of 1e-5, batch size of 512, and prompt dropout rate of 0.1\footnote{Prompts with more than 77 tokens (CLIP's upper limit) were dropped. This happened for <1\% of the data in each training set.}. We fine-tuned the model training both UNet and CLIP weights and used a 50\%-50\% mixture of RECAP Short and RECAP Long captions (RECAP Mix) as this performed best. The results in the main text are all from this configuration. 

RECAP is independent of the sampling method, so we can use any sampling method with it. That said, for all of the experiments in this work, we used DDIM sampling with 50 inference steps and a guidance scale of 7.5.

\section{Results} \label{results}
We compare the RECAP model to two models: Baseline and Alttext. Baseline is the Stable Diffusion v1.4 model. Alttext is the baseline model fine-tuned for the same number of steps and on the same set of images as RECAP, but with the original captions (Alttext) instead of the RECAP captions. The Alttext model resolves contamination concerns, as it includes the exact set of images. In all the comparisons we used the same random seeds across models.

We compare the models using a variety of automated metrics, human evaluation and a qualitative evaluation of examples. We observe improvements in all metrics (see \cref{table:tise_main,table:human_eval}).

\subsection{Automated metrics}
We evaluated the performance and semantic capabilities of the RECAP model using a battery of metrics suggested by \cite{dinh2022tise} (using their publicly available code) on the MS-COCO validation dataset. \cref{table:tise_main} contains a summary of the results.

To assess overall generation quality we use the standard FID metric and observe that images generated with the RECAP model have a significantly better score ($17.87\rightarrow14.84$).

In addition, we assess the semantic capabilities of our model: to check that the model generates faithfully the requested objects we use Semantic Object Accuracy ($80.80\rightarrow86.17$), to check the number of generated objects we use Counting Alignment errors ($1.44\rightarrow1.32$), to check that the locations of objects are correct we use Positional Alignment ($57.60\rightarrow62.42$), and finally to check the overall adherence to the prompt we measure Clip score ($92.78\rightarrow93.80$).

In all the metrics, we see no improvement in the Alttext model compared to the baseline, proving the improvements stem from the captions themselves and not from the additional training.

Note that following \cite{dinh2022tise}, we omit the IS and O-IS metrics, as all Stable Diffusion models yield a higher score than real images. As noted by \cite{dinh2022tise}, IS is better suited for datasets with a single object and does not perform well for the MS-COCO dataset, containing multiple objects per each image. 

Additional details on the automated metrics are in  \cref{fid,object_accuracy,counting,text_relevance}.

\subsection{Human evaluation}
For complementary evaluation of model performance, we used human raters. Results are summarized in \cref{table:human_eval}.

Raters were asked to select images generated from each model, only if they successfully follow a given prompt. We evaluated once on 200 random prompts from the MS-COCO validation set, and separately on the challenging DrawBench dataset \cite{Imagen}. We presented four images (from different seeds) for each prompt, using the same seeds across models.

We calculated two metrics: percentage of successful image generation across all prompts and seeds (i.e. given a prompt and a seed, the chance of a successful image generation); and percentage of at least one successful image generation for a given prompt, out of four seeds (i.e. given a prompt, the chance of successfully generating an image for it, in four attempts).
We see a relative 64.3\% improvement in successful image generation on MS-COCO, and a 41.7\% improvement on DrawBench. We also see a relative improvement of 42.1\% in successful prompt generation on MS-COCO and 37.5\% improvement on DrawBench. The Alttext model showed minor improvement on the MS-COCO dataset (12\%-13\%) and did not improve the DrawBench dataset. Further details can be found in \cref{human_eval}.

\begin{table*}
    \centering
    \begin{tabular}{c c c c c c c c} 
         \hline
         & FID$\downarrow$ & O-FID$\downarrow$& SOA-C$\uparrow$ & SOA-I$\uparrow$ & CA$\downarrow$ & PA$\uparrow$  & RP$\uparrow$ \\
         \hline
        Baseline & 17.87 & 8.19 & 78.90 & 80.80 & 1.44 & 57.60 & 92.78 \\
        Alttext & 17.53 & 8.90 & 78.99 & 80.85 & 1.47 & 57.40 & 91.32 \\
        RECAP & \textbf{14.84} & \textbf{6.23} & \textbf{84.34} & \textbf{86.17} & \textbf{1.32} & \textbf{62.42} & \textbf {93.80} \\
         \hline
        Real Images & 2.62 & 0.00 & 90.02 & 91.19 & 1.05 & 100.0 & 83.54 \\
         \hline
    \end{tabular}
    \caption{Results for the automated metrics for RECAP model vs. baseline and Alttext models. RECAP model improves across all metrics.}
    \label{table:tise_main}
\end{table*}

\begin{table*}
    \centering
    \begin{tabular}{c c c c c} 
         \hline
          & MS-COCO & MS-COCO & DrawBench & DrawBench  \\
          & Successful Images & Successful Prompts & Successful Images & Successful Prompts \\
         \hline
         Baseline & 29.3\% & 53.5\% & 15.6\% & 33.1\% \\
         Alttext & 33.4\% & 60.0\% & 13.2\% & 33.8\% \\
         RECAP & \textbf{48.1\%} & \textbf{76.0\%} & \textbf{22.1\%} & \textbf{45.5\%} \\
         \hline
    \end{tabular}
    \caption{Human evaluation results comparing RECAP, Alttext and Baseline models, on two benchmarks (MS-COCO and DrawBench) for two metrics: percentage of images generated that fully follow the prompt, and percentage of prompts with at least one generated image (out of four seeds) that fully followed the prompt.}
    \label{table:human_eval}
\end{table*}

\subsection{Qualitative Results}
\cref{fig:top_example} provides representative examples where RECAP outperforms the base Stable Diffusion model (and sometimes also larger models, see \cref{appendix_other_models}). 

Generally, we observe that RECAP can better interpret relations between entities. Prepositions like "near", "through", "made off" are often ignored by the base Stable Diffusion model. The RECAP model applies the prepositions correctly, while the base model often resorts to the most common relation between the entities (based on the training data distribution). For example, in the prompt "\emph{A pizza near a pineapple}", the base model places the pineapple on the pizza, as the probable relation between the two, while RECAP generates a pineapple near it, as requested.

RECAP also better handles cases where different modifiers are applied to multiple entities (e.g. "\emph{A red bench and a yellow clock}"). The base model will treat the sentence as a bag of words, applying all modifiers to all entities or ignore some of them. RECAP is also able to interpret complex modifiers like anaphors. For examples, in the prompt "\emph{Two flowers, one is blue and the other one is green}" it  understands that "one" refers to a flower. 

\section{Analysis} \label{analysis}
We hypothesize that the underlying improvement in image generation quality, as measured in the results above, stems from two improvements in the training captions: (1) reducing the discrepancy between the train and inference prompts, and (2) giving the model more textual information per image, thus improving the training sample efficiency.
Below we provide an ablation analysis showing that RECAP captions achieve both properties, and that the resulting model benefits from both.

\subsection{Comparing Different Caption Types}
\label{ablation_caption_distribution}

\subsubsection{Generated Captions} \label{ablation_caption_distribution_captions}
\cref{table:textstat} compares the generated captions in the training set, to the caption in the MS-COCO validation set, on a variety of language metrics. We used MS-COCO to represent the prompts we expect to see in inference time. We observe that our generated captions are closer in distribution to MS-COCO in several senses. 

First, we compare them by using standard language metrics from the \emph{textstat} python package\footnote{\url{https://pypi.org/project/textstat}} calculated over the 10M examples in each train dataset.
\emph{Flesch Reading Ease} score is a standard measure for how difficult it is to read a text in a given language. We observe that RECAP is able to generate easy to read sentences, while the original Alttext is often difficult to read. Similarly, the \emph{text\_standard} score is a consensus score based on a battery of metrics to estimate the smallest grade of a native speaking reader to be able to read the text. RECAP generates texts which a 4th grader (the lowest possible) can read, while the original Alttext is estimated to require 8th graders to fully understand.

To measure more directly the reduction in train-inference skew, we compare the distribution of the embedded texts. Since Stable Diffusion is using CLIP to encode the text, we calculate the Fréchet distance between the CLIP embeddings of the various datasets\footnote{Similarly to the FID metric used to compare between image distributions, we take the Fréchet distance of the CLIP embedding distributions, modeled as a Gaussian.}. Results are summarized in \cref{table:textstat}. As suspected, RECAP generated captions are closer in distribution to the MS-COCO captions than the Alttext captions. Furthermore, the RECAP Short captions are closer to MS-COCO than RECAP Long captions. These results are in line with the improvement in FID of the images from models trained with the corresponding datasets, as detailed in \cref{ablation_caption_distribution_images}.

Next, we measured how well the automatically generated captions describe the images, using a human evaluation of 100 random images, asking raters to score each caption considering both faithfulness to the image, and completeness in describing the image, on a scale of 1-5. RECAP generated captions were rated as more faithful and complete (with an average score of 3.58 for RECAP Short and 4.3 for RECAP Long) than Alttext captions (scoring on average 2.9). This supports our hypothesis that automated captions make training more efficient by providing more textual information. 

\begin{table*}
    \centering
    \begin{tabular}{c c c c c} 
         \hline
         & Alttext & RECAP Long & RECAP Short & MS-COCO\\
         \hline
        \# sentences & 1.15 & 2.15 & 1.00 & 1.00 \\
        \# words & 11.38 & 21.29 & 5.89 & 10.45 \\
        \# letters & 61.08 & 87.96 & 24.39 & 41.67 \\
        Flesch Reading Ease score & 45.71 (= Difficult) & 88.35 (= Easy) & 86.61 (= Easy) & 86.93 (= Easy) \\
        \emph{text\_standard} score & 8.78 (8th grader) & 4.50 (4th grader) & 3.11 (4th grader) & 4.59 (4th grader) \\
        Distance to MS-COCO & 0.45 & 0.24 & 0.18 & 0.00 \\
         \hline
    \end{tabular}
    \caption{Comparison of average statistics on the captions generated by RECAP vs. the original Alttext over the 10m examples in the training set, and the 10k examples in the MS-COCO validation set. Distance to MS-COCO is the Fréchet distance of the CLIP embeddings of the RECAP caption sets (calculated on 250k samples) vs. the MS-COCO validation set. Note that since the embeddings are normalized, the distance in between 0-1.}
    \label{table:textstat}
\end{table*}

\subsubsection{Generated Images} \label{ablation_caption_distribution_images}
\begin{figure*}[htbp]
    \includegraphics[width=0.5\linewidth]{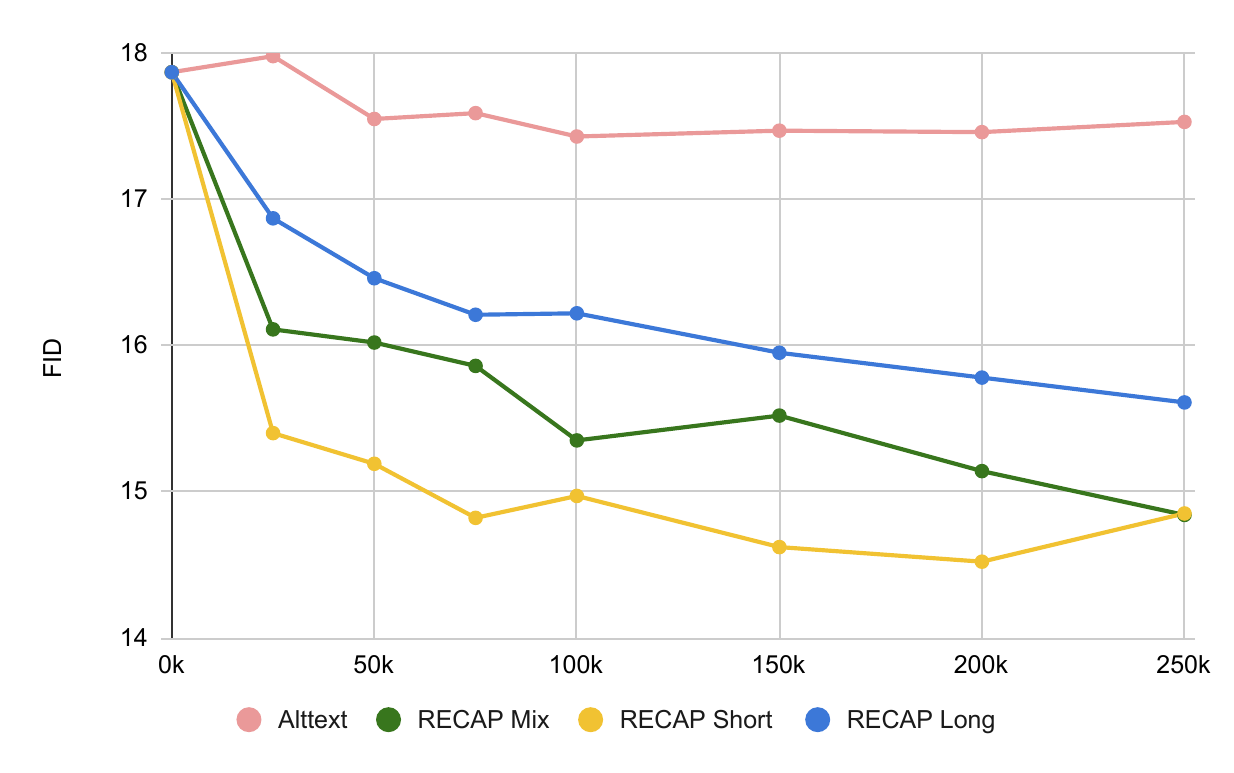}
    \includegraphics[width=0.5\linewidth]{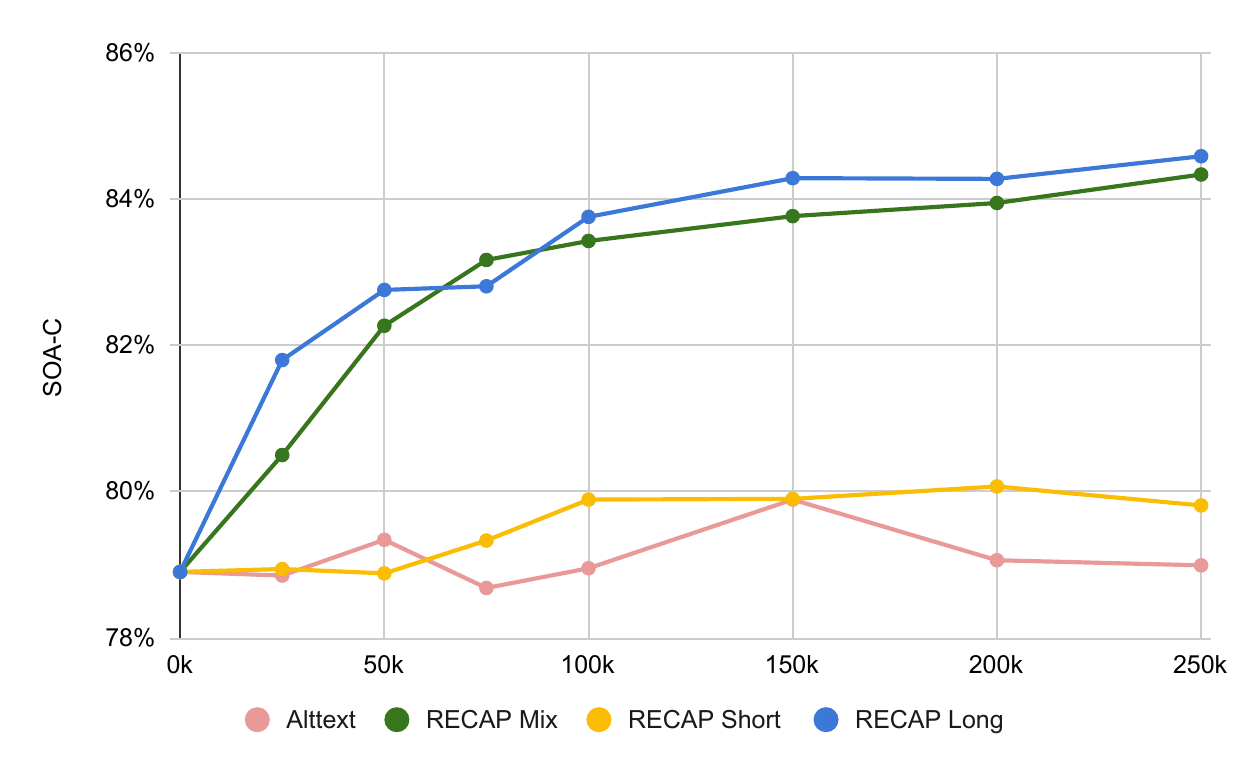}
    \captionof{figure}{FID (left) and SOA (right) scores for various checkpoints of the Alttext model vs. RECAP models, comparing models that trained on different sets of captions. 0 fine-tuning steps is vanilla Stable Diffusion 1.4. Lower FID is better. We see that RECAP Short achieves better FID and faster, but with no semantic improvement. RECAP Long achieves some FID improvement with significant semantic improvement, and RECAP Mix achieves both.}
    \label{fig:fid_and_soa_captions}
\end{figure*}

Next, we compared the results of fine-tuning Stable Diffusion on our two caption types, RECAP Long and RECAP Short. The results are summarized in \cref{table:tise_captions}. We observe that training on RECAP Short captions achieves better FID scores, and faster, but with little semantic improvement, while the RECAP Long captions exhibit significant semantic improvement (see representative metrics in \cref{fig:fid_and_soa_captions}). Mixing the caption sets (RECAP Mix) provides the best of both worlds.

Note that the improvement in FID on the MS-COCO validation set, is correlated to the improvement in the Fréchet distance of the generated captions, as detailed in \cref{ablation_caption_distribution_captions}. This indicates that the semantic improvement in the RECAP Long model stems from improved training sample efficiency, and not only from reducing the train-inference skew.

\begin{table*}
    \centering
    \begin{tabular}{c c c c c c c c} 
         \hline
         & FID$\downarrow$ & O-FID$\downarrow$ & SOA-C$\uparrow$ & SOA-I$\uparrow$ & CA$\downarrow$ & PA$\uparrow$ & RP$\uparrow$ \\
         \hline
        Baseline & 17.87 & 8.19 & 78.90 & 80.80 & 1.44 & 57.60 & 92.78 \\
        Alttext & 17.53 & 8.90 & 78.99 & 80.85 & 1.47 & 57.40 & 91.32 \\
        RECAP Short & 14.85 & \textbf{5.81} & 79.81 & 81.95 & 1.43 & 60.00 & 93.30 \\
        RECAP Long & 15.61 & 7.48 & \textbf{84.59} & 86.16 & 1.32 & \textbf{63.27} & 93.26 \\
        RECAP Mix & \textbf{14.84} & 6.23 & 84.34 & \textbf{86.17} & \textbf{1.32} & 62.42 & \textbf{93.80} \\
         \hline
        Real Images & 2.62 & 0.00 & 90.02 & 91.19 & 1.05 & 100.0 & 83.54 \\
         \hline
    \end{tabular}
    \caption{Results for the automated metrics for RECAP models vs. baseline and Alttext models, comparing models that trained on different sets of captions.}
    \label{table:tise_captions}
\end{table*}

\subsection{Training Different Model Weights} \label{analysis_weights}
To further explore the contribution of the improved examples to each part of the Stable Diffusion model, we compared training only the UNet weights, vs. the CLIP weights, vs. both, using the same training and evaluation procedures. For simplicity, we only report the results for RECAP Mix vs. Alttext and baseline models.

The results are summarized in \cref{table:tise_weights}. Overall, as expected, training more weights achieves better performance. Interestingly, training only CLIP weights (which are $\backsim12\%$ of the total weights) achieves better FID, with less training steps, but with little semantic improvement. Training both CLIP and UNet weights results in significantly higher improvement to semantic scores than training only on one of them. See \cref{appendix_weights} for more details.

We believe that training CLIP weights mainly reduces the skew in the distribution of texts between the training set and the evaluation set, while training the UNet weights mostly improves the alignment of the text to the image.

\begin{table*}
    \centering
    \begin{tabular}{c c c c c c c c} 
         \hline
         & FID$\downarrow$ & O-FID$\downarrow$ & SOA-C$\uparrow$ & SOA-I$\uparrow$ & CA$\downarrow$ & PA$\uparrow$ & RP$\uparrow$ \\
         \hline
        Baseline & 17.87 & 8.19 & 78.90 & 80.80 & 1.44 & 57.60 & 92.78 \\
        Alttext UNet & 16.83 & 8.20 & 76.90 & 78.72 & 1.50 & 56.05 & 91.89 \\
        Alttext CLIP & 17.04 & 8.32 & 79.08 & 81.10 & 1.52 & 58.72 & 91.34 \\
        Alttext UNet+CLIP & 17.53 & 8.90 & 78.99 & 80.85 & 1.47 & 57.40 & 91.32 \\
        RECAP Mix UNet & 15.49 & 6.47 & 82.15 & 84.04 & 1.40 & 59.77 & 92.76 \\
        RECAP Mix CLIP & \textbf{14.60} & \textbf{6.09} & 80.19 & 82.04 & 1.37 & 60.44 & 92.48 \\
        RECAP Mix UNet+CLIP & 14.84 & 6.23 & \textbf{84.34} & \textbf{86.17} & \textbf{1.32} & \textbf{62.42} & \textbf{93.80} \\
         \hline
        Real Images & 2.62 & 0.00 & 90.02 & 91.19 & 1.05 & 100.0 & 83.54 \\
         \hline
    \end{tabular}
    \caption{Results for the automated metrics for RECAP models vs. baseline and Alttext models, comparing models that trained different set of weights.}
    \label{table:tise_weights}
\end{table*}

\section{Summary and Discussion}
In this paper, we show how text-to-image models can be improved across the board by training on synthetically generated captions. We performed an in-depth analysis demonstrating that short descriptions which narrow the train-inference gap are helpful, as are long and detailed descriptions that improve sample efficiency despite being different from the inference set. We further demonstrated that mixing these descriptions in the training set improves all fronts simultaneously. 

There are several interesting directions for future work. It would be interesting to check if by tuning the captioning model to produce ample detail in narrow areas, the same recipe can be used to improve semantic capabilities in new domains (for example, could we create models that can accurately generate hair styles, room designs, facial expressions, clothing, etc. based on detailed descriptions?). Similarly, it is possible to use RECAP to train T2I models in domains that lack textual captions altogether (e.g., a personal photo album or screencaps of TV shows). We conducted initial experiments here, and they show a lot of promise.

We experimented with fine-tuning a model with the RECAP captions, but it would be interesting to compare that to a model that was pre-trained on the RECAP captions from scratch. Relatedly, it would be interesting to further experiment with different mixtures of the three caption types we have (RECAP Short, RECAP Long, and Alttext). Even more generally, we could imagine creating and mixing together several more flavors of recaptioning models. This could allow us to circumvent the token limit (by training on the same image multiple times with a different subset of the caption each time). Regardless of token limit, it could be interesting to explore training on several shorter captions per image instead of a single long one. It would also be interesting to explore the effects of RECAP on larger models trained on larger datasets. 

Finally, RECAP shows the importance of high quality datasets, and that it can be improved with synthetic data, we hope that this provides yet another encouragement to apply such techniques even beyond the T2I domain.

\section{Acknowledgments}
We would like to extend our gratitude to Eyal Molad, Matan Kalman, Jason Baldridge, and the Theta Labs team at Google, for great reviews, suggestions, and support to this paper.
\clearpage
\clearpage

{
\bibliography{references}
}

\clearpage

\appendix 

\section{Detailed Results} \label{appendix_results}

The following sub-sections explain in more detail the metrics used to evaluate our model, as well as provides plots of several metrics over the number of training steps.

\subsection{Image Realism} \label{fid}
FID \cite{NIPS2017_8a1d6947} is a measure of how close two image datasets are in terms of the distribution of the semantic content across a large set of photos in each dataset. It is commonly used for evaluation of text-to-image models, by taking a paired set of texts and images, and comparing the set of original images to a set of newly generated images from the texts. A lower score means the distributions are more alike. \newline
O-FID \cite{dinh2022tise} is a variant of FID where off-the-shelf object detector crops out objects from the images first, and FID is calculated on the cropped image sets, giving somewhat more granular measurement of the distribution of objects.

\begin{figure*}[htbp]
    \includegraphics[width=0.5\linewidth]{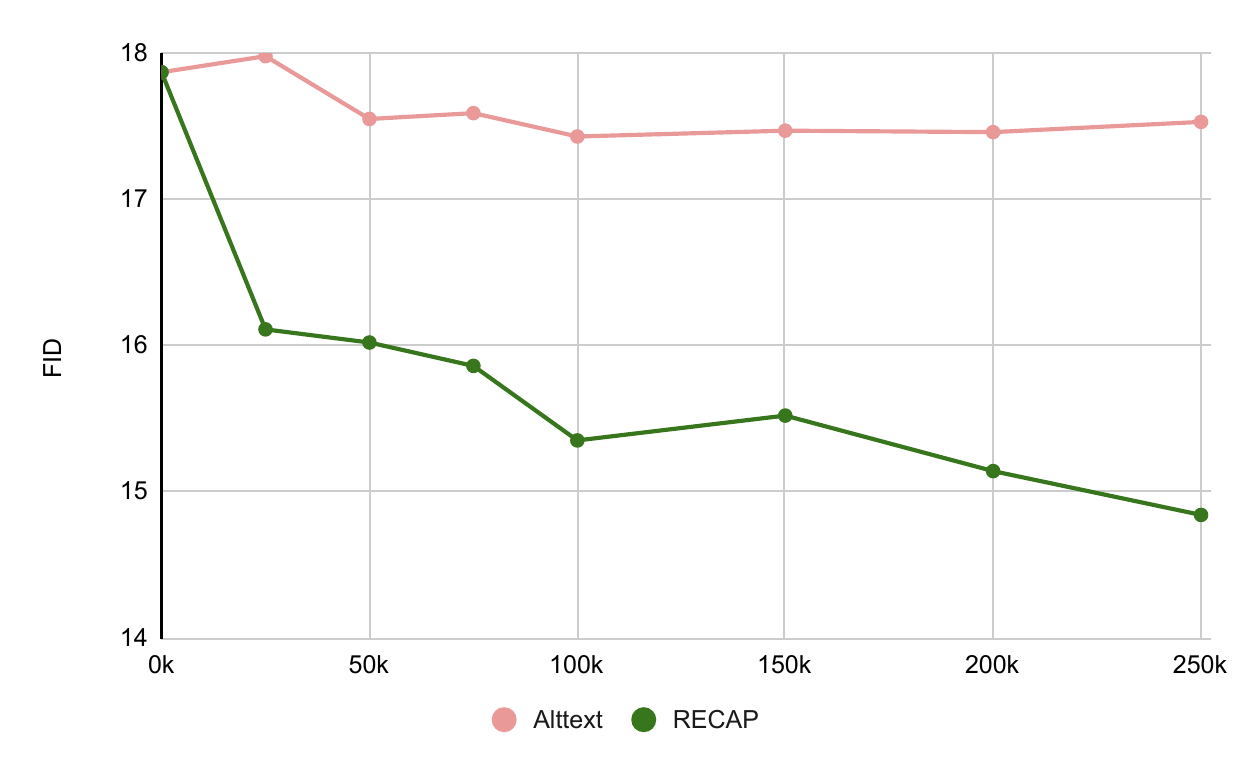}
    \includegraphics[width=0.5\linewidth]{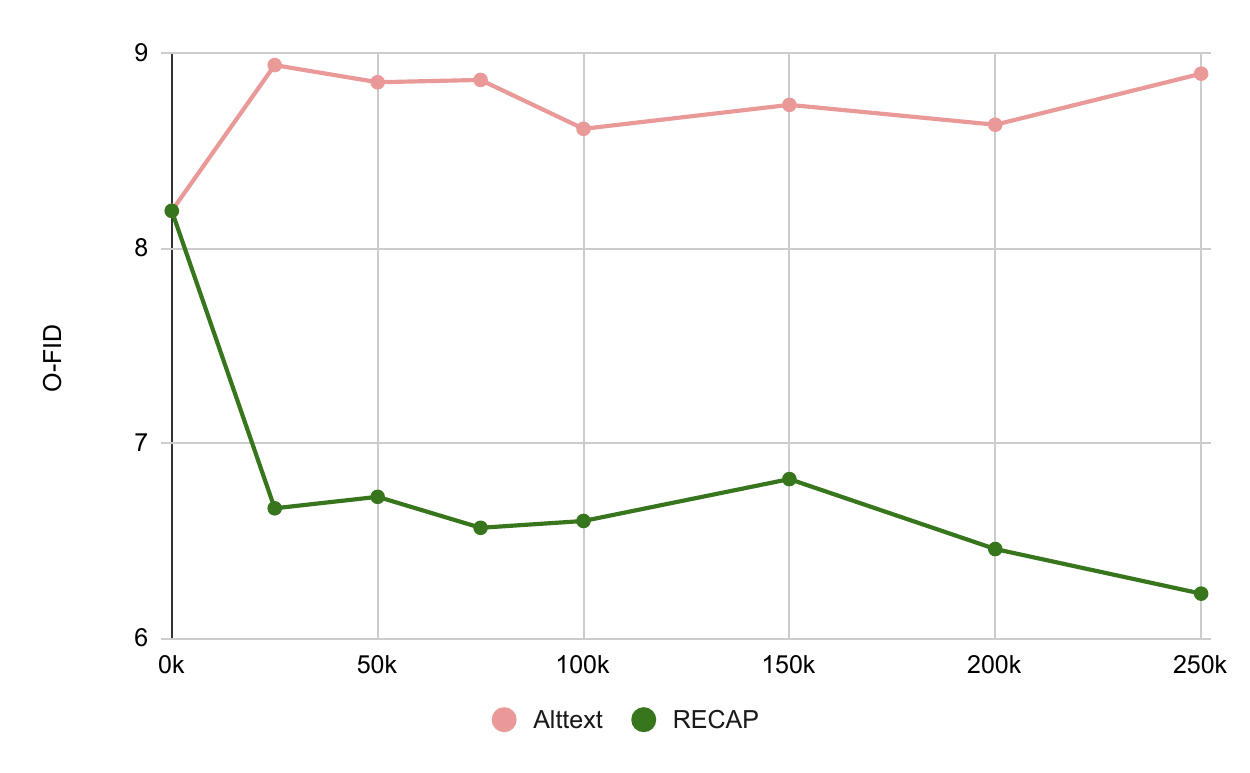}
    \captionof{figure}{FID (left) and O-FID (right) scores for various checkpoints of the Alttext model vs. the RECAP model. 0 fine-tuning steps is vanilla Stable Diffusion 1.4. Lower score is better.}
    \label{fig:fid_simple}
\end{figure*}

\cref{fig:fid_simple} shows the calculated FID and O-FID scores for several checkpoints throughout training. It is clear that the RECAP model improves significantly the FID and O-FID scores over the MS-COCO dataset, while fine-tuning the same amount of steps on the same set of images, but with the original Alttext captions, produces only a minor improvement.

\subsection{Semantic Object Accuracy} \label{object_accuracy}
The SOA metric was suggested in \cite{Hinz_2022} and is used to evaluate the accuracy of a text-to-image generative model, by measuring how well it follows the instructions to generate specific objects as part of a prompt. To do so, it uses off-the-shelf specialized object detectors of 80 different classes (e.g. a motorcycle or a keyboard) on labeled data. There are two variants to this metric, one that averages across images (SOA-I) and another that averages across classes (SOA-C).

\begin{figure*}[htbp]
    \includegraphics[width=0.5\linewidth]{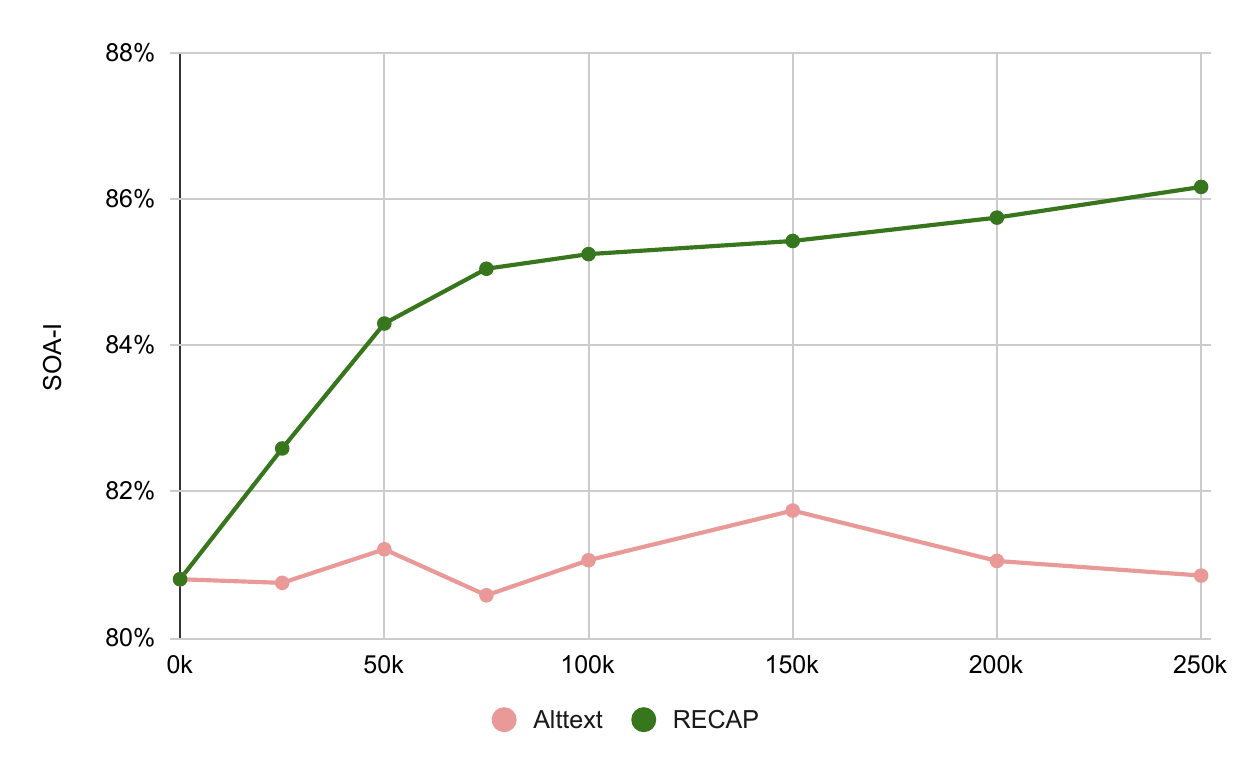}
    \includegraphics[width=0.5\linewidth]{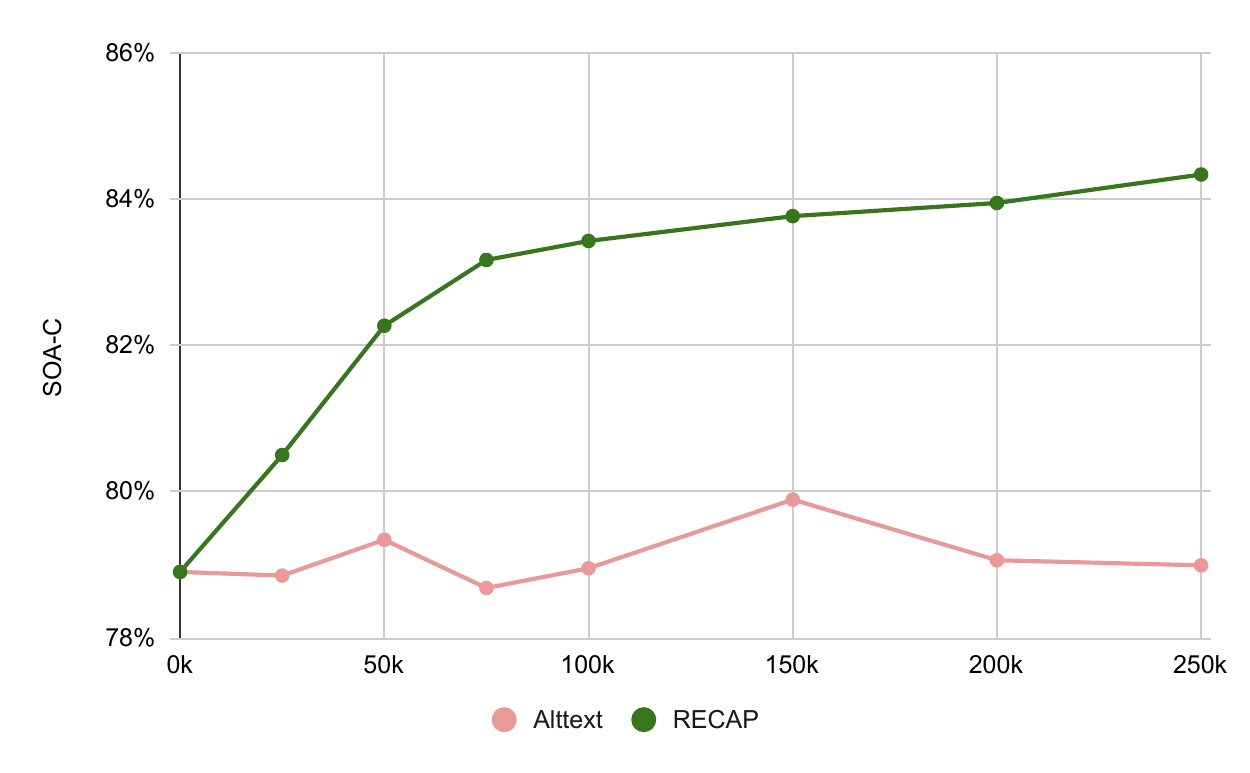}
    \captionof{figure}{Semantic Object Accuracy (SOA) scores for various checkpoints of the Alttext model vs. the RECAP model. 0 fine-tuning steps is vanilla Stable Diffusion 1.4. SOA-I (left) averages across images, while SOA-C (right) averages across classes.}
    \label{fig:soa_simple}
\end{figure*}

Results can be found in \cref{fig:soa_simple}. We see a very significant improvement for the RECAP model vs. the baseline, while the Alttext model does not show any improvement.

\subsection{Counting and Positional Alignments} \label{counting}
Generative text to image models are known to struggle counting objects, i.e. if a prompt specifies a specific number of objects (e.g. “3 birds”) the model often generates a different number of objects. In the Counting Alignment (CA) metric \cite{dinh2022tise}, MS-COCO was filtered to prompts which contain a specific instruction to generate a known number of objects. An off-the-shelf counting model was used to count the number of expected objects in the generated image (per object type). The lower the score the better. 

Similarly, generative text to image models often struggle to follow positional cues, e.g. “a girl in front of a boy” or “a plate of avocado under the table”. In the Positional Alignment (PA) metric \cite{dinh2022tise}, MS-COCO was filtered to prompts with specific positional cues. Each image generated by such prompt gets a CLIP score vs. the original prompt, and also vs. each replacement of the positional cue in the original prompt with a different (wrong) one (e.g. "under" instead of "in front of").

\begin{figure*}[htbp]
    \includegraphics[width=0.5\linewidth]{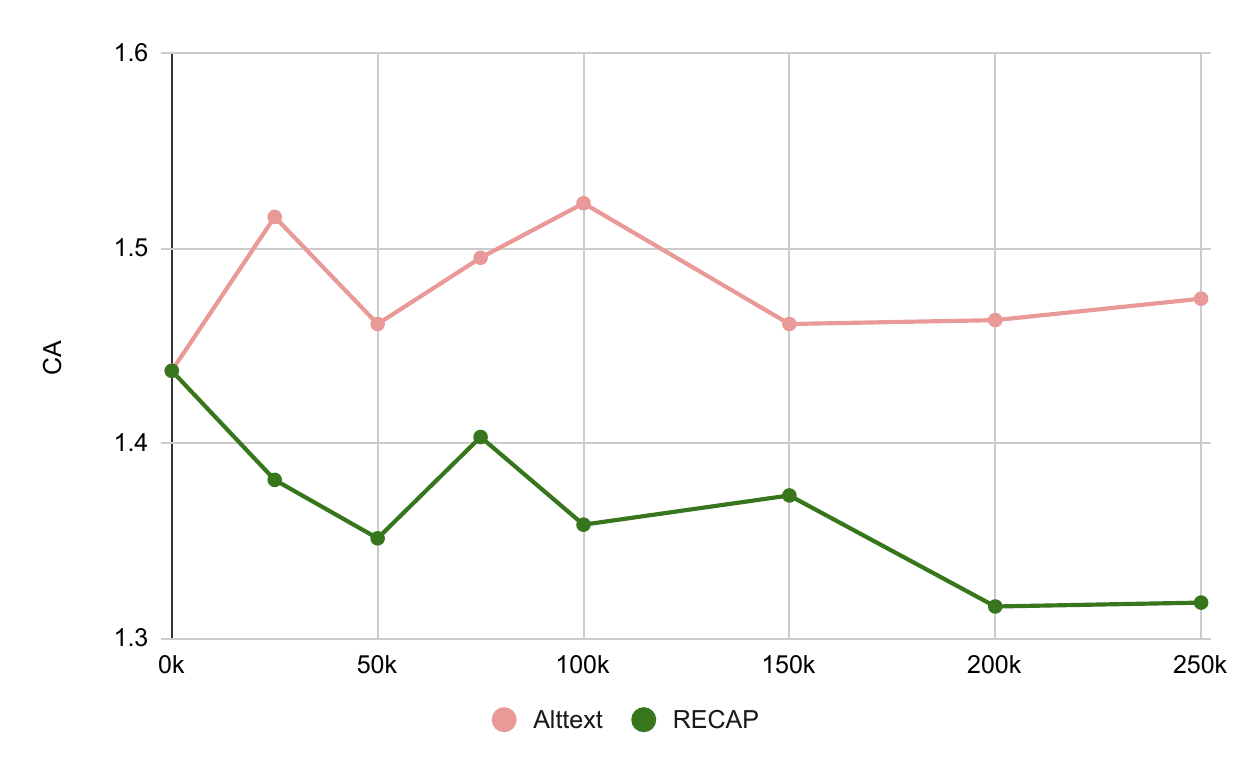}
    \includegraphics[width=0.5\linewidth]{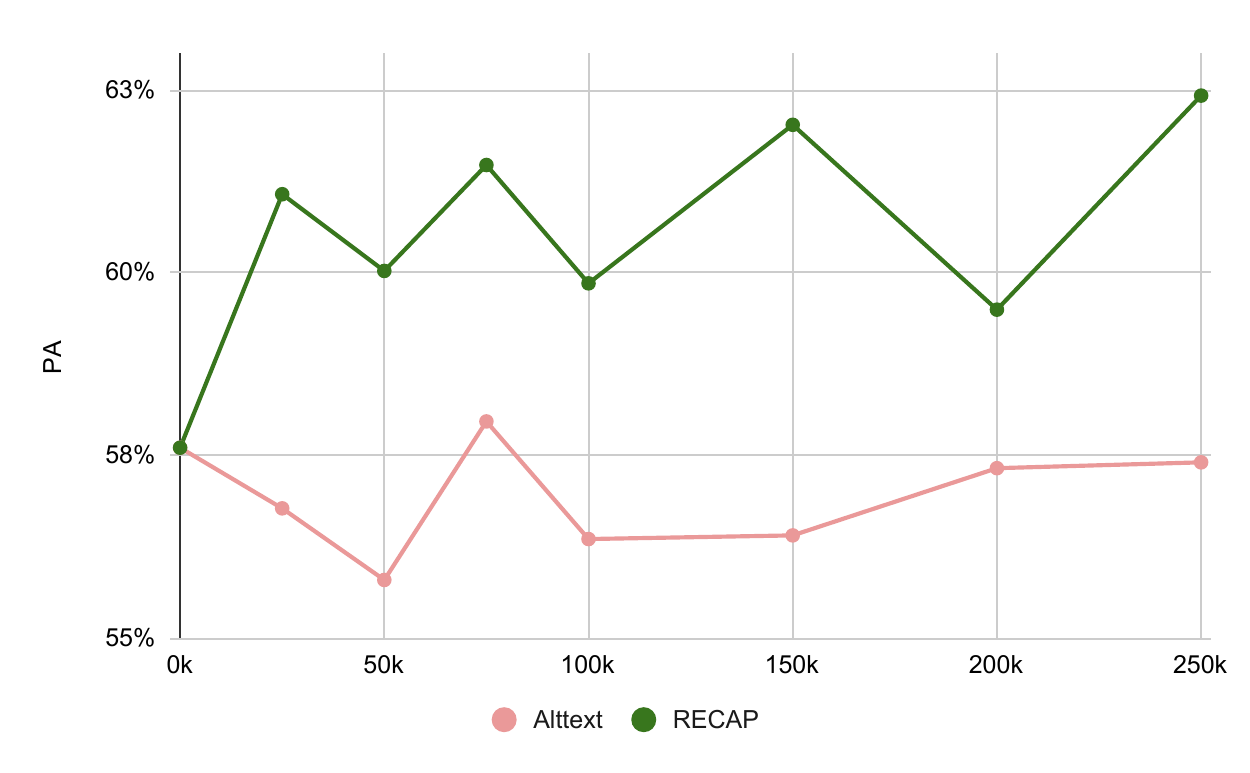}
    \captionof{figure}{Counting Alignment (CA) (left) and Positional Alignment (PA) (right) scores for various checkpoints of the Alttext model vs. the RECAP model. 0 fine-tuning steps is vanilla Stable Diffusion 1.4. Lower CA score is better.}
    \label{fig:ca_pa_simple}
\end{figure*}

Results for both metrics can be found in \cref{fig:ca_pa_simple}. Once again, the RECAP model improves the baseline while the Alttext model does not.

\subsection{Text Alignment} \label{text_relevance}
R-Precision metric (RP), also known as CLIP score, \cite{8578241} is a popular measure to how close a prompt is to an image generated from it, by using the CLIP embeddings distance between each prompt and the image generated from it. However, CLIP is also used by Stable Diffusion as the text encoder, producing a bias towards images generated by Stable Diffusion model, and in particular yielding higher RP score for the generated images vs. the original real images. Still, we see that relative to the base model (scoring $92.78$), the Alttext model achieves lower score ($91.31$, probably due to the small dataset size which reduces the training data variance), while the RECAP model improves it ($93.8$, despite the small dataset size).

\subsection{Human Evaluation} \label{human_eval}
We sent samples from the base, Alttext and RECAP models for human evaluation. Raters were presented with four images for each model along with the prompt used to generate the image. The instructions were to only select images that strictly followed the prompt and did not contain any major deformities (minor deformities are fairly common with Stable Diffusion 1.4). Presented results are averaged across raters.

We evaluate all of these models on 200 randomly sampled prompts from MS-COCO and the DrawBench dataset \cite{Imagen}\footnote{2 prompts which were >77 CLIP tokens were dropped}, which Stable Diffusion 1.4 is known to have difficulty with. The RECAP model generates 64.3\% ($29.25\%\rightarrow48.06\%$) more valid images, and is 42.1\% ($53.5\%\rightarrow76\%$) more likely to be able to generate at least one valid image (out of four seeds) over the base model. This indicates that RECAP is both better at generating images as well as being capable of following more difficult prompts (\cref{table:human_eval}).

\subsection{Model Weights} \label{appendix_weights}
\begin{figure*}[htbp]
    \includegraphics[width=0.5\linewidth]{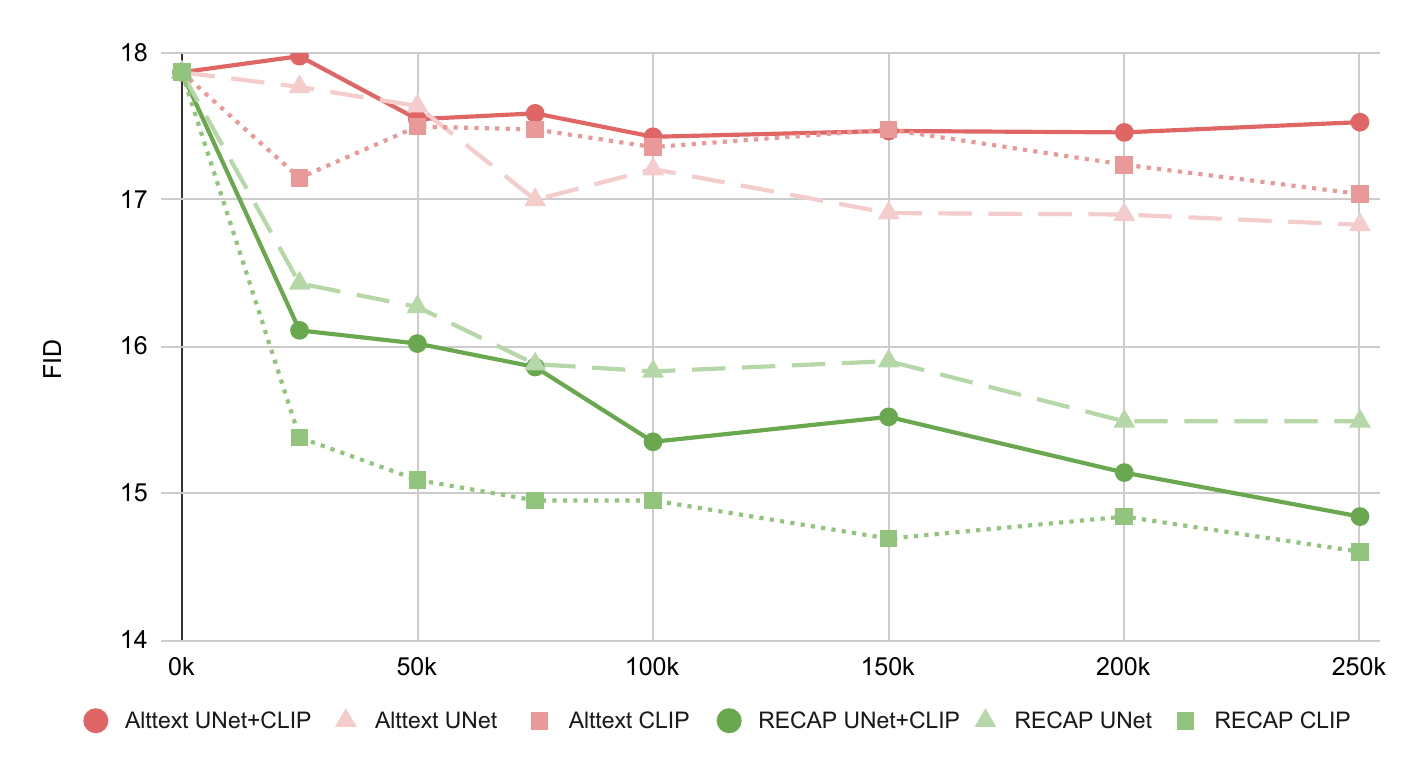}
    \includegraphics[width=0.5\linewidth]{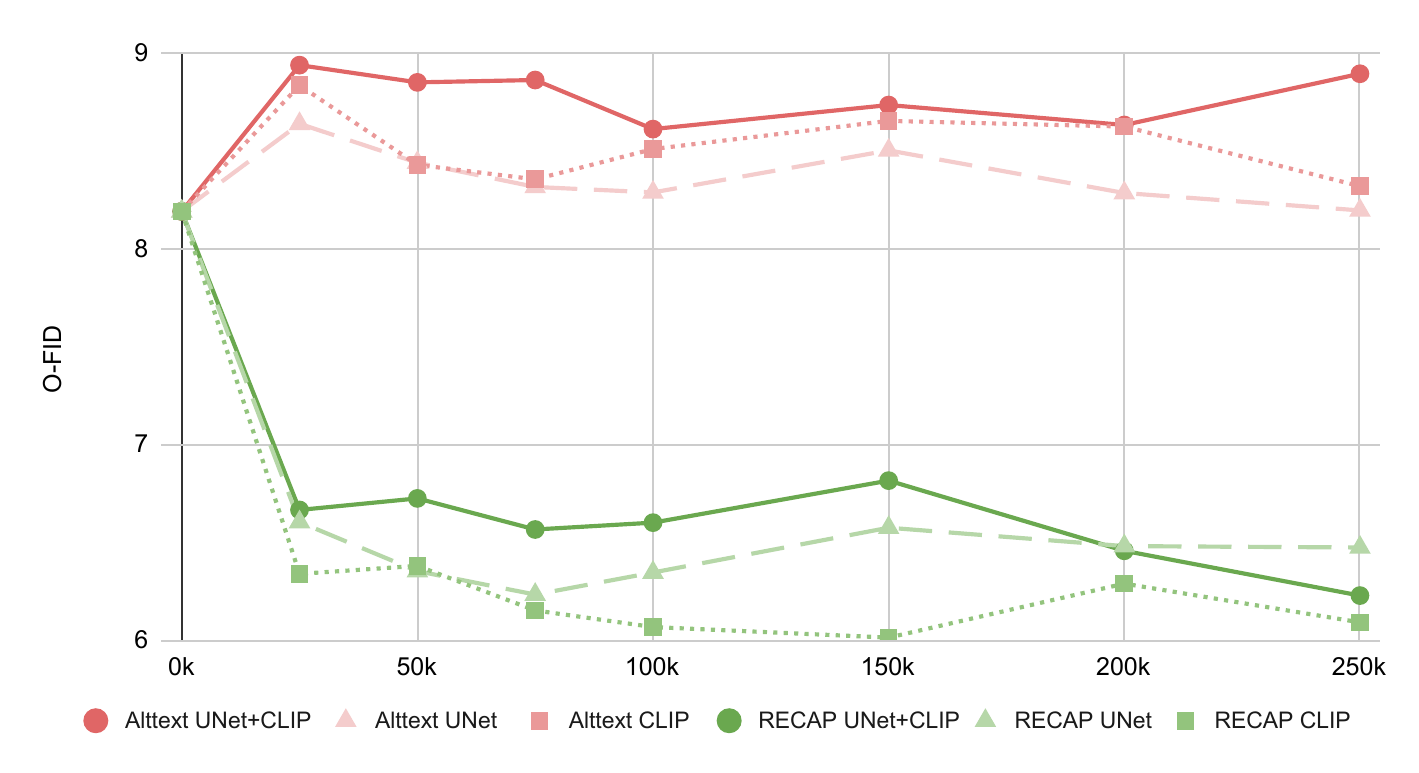}
    \captionof{figure}{FID (left) and O-FID (right) scores for various checkpoints of the Alttext model vs. RECAP models, comparing models that trained on different set of weights. 0 fine-tuning steps is vanilla Stable Diffusion 1.4. Lower is better.}
    \label{fig:fid_weights}
\end{figure*}

\begin{figure*}[htbp]
    \includegraphics[width=0.5\linewidth]{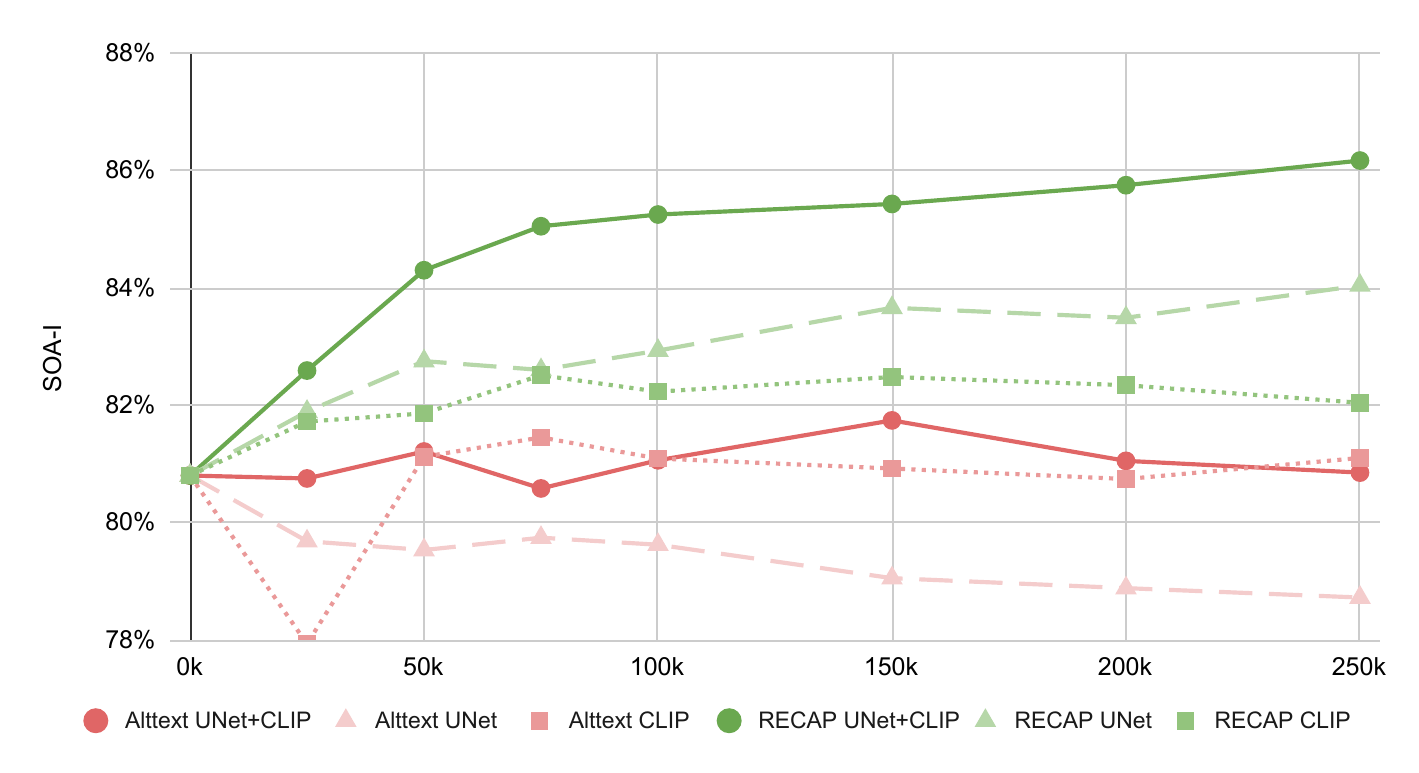}
    \includegraphics[width=0.5\linewidth]{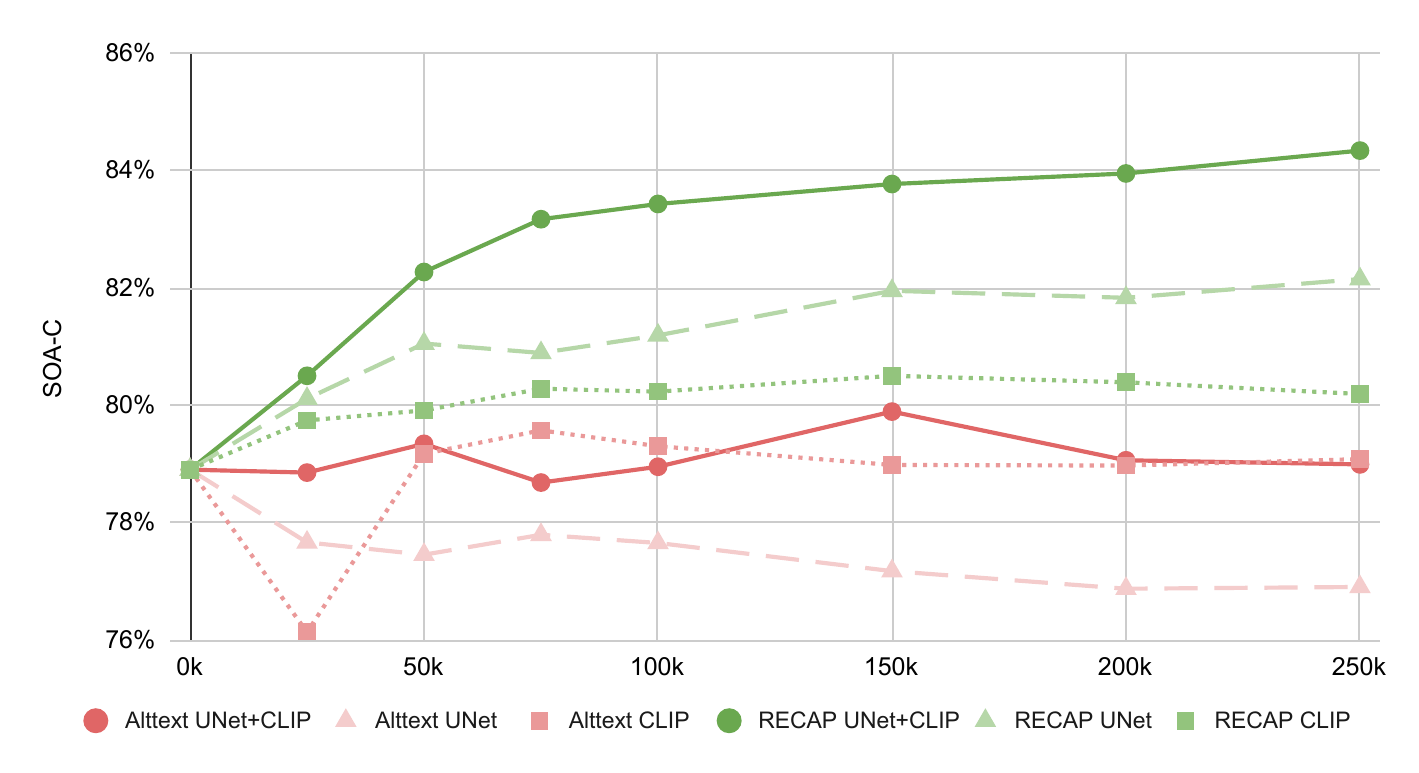}
    \captionof{figure}{Semantic Object Accuracy (SOA) scores for various checkpoints of the Alttext model vs. RECAP models, comparing models that trained on different set of weights. 0 fine-tuning steps is vanilla Stable Diffusion 1.4. SOA-I (left) averages across images, while SOA-C (right) averages across classes.}
    \label{fig:soa_weights}
\end{figure*}

We provide additional plots for the various models trained with different subset of unfrozen weights, as described in \cref{analysis_weights}, in \cref{fig:fid_weights,fig:soa_weights}.

\twocolumn[{
\section{Other Image Generation Models} \label{appendix_other_models}

\cref{fig:midjourney_examples} shows example images generated by SDXL and Midjourney for the prompts in the top figure.

\begin{center}
    \includegraphics[width=1.0\linewidth]{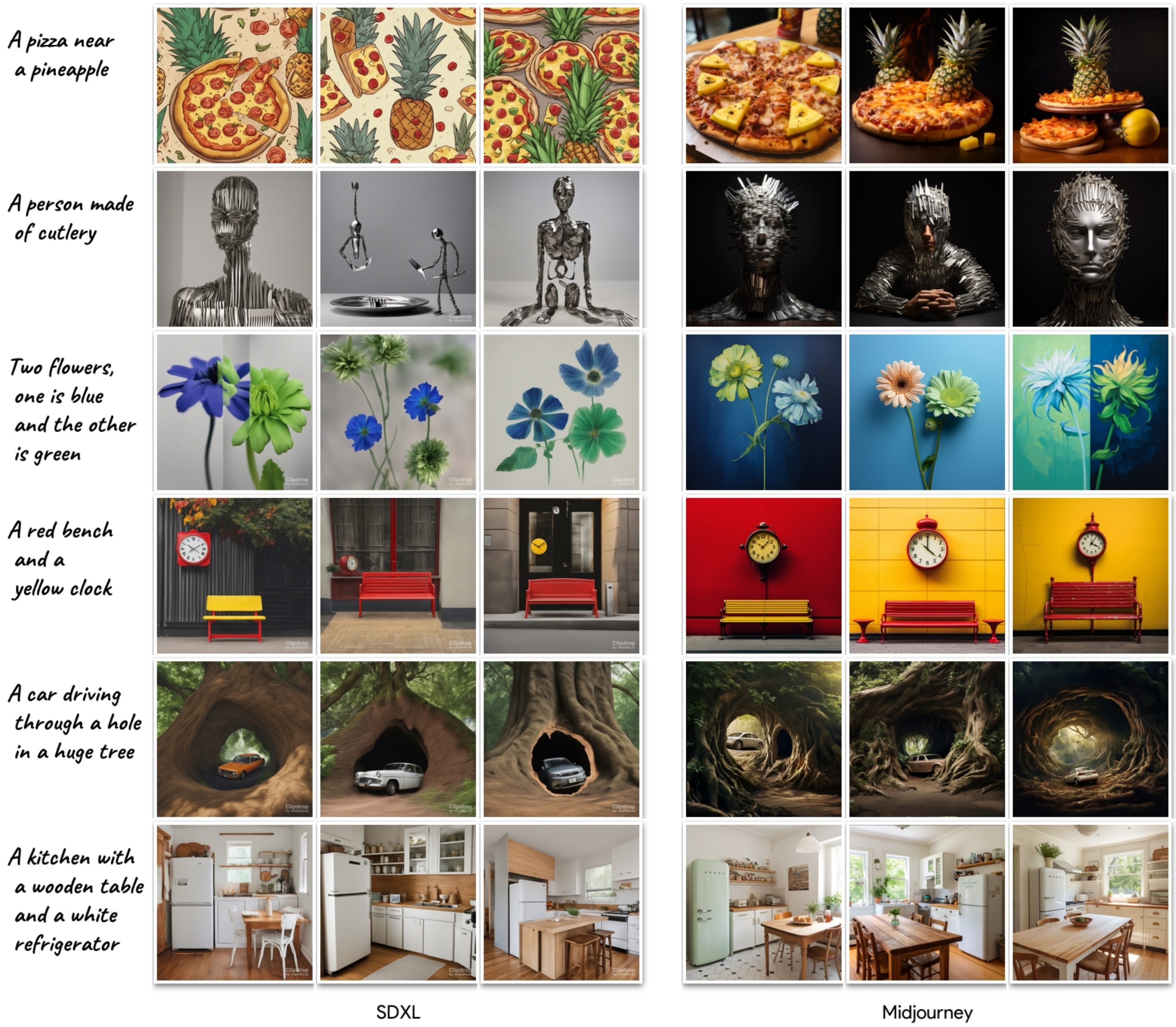}
\captionof{figure}{Examples of images generated by SDXL 1.0 and Midjourney 5.1 for the prompts in the top figure.}
    \label{fig:midjourney_examples}
\end{center}
}]

\begin{figure*}
    \includegraphics[width=0.88\linewidth]{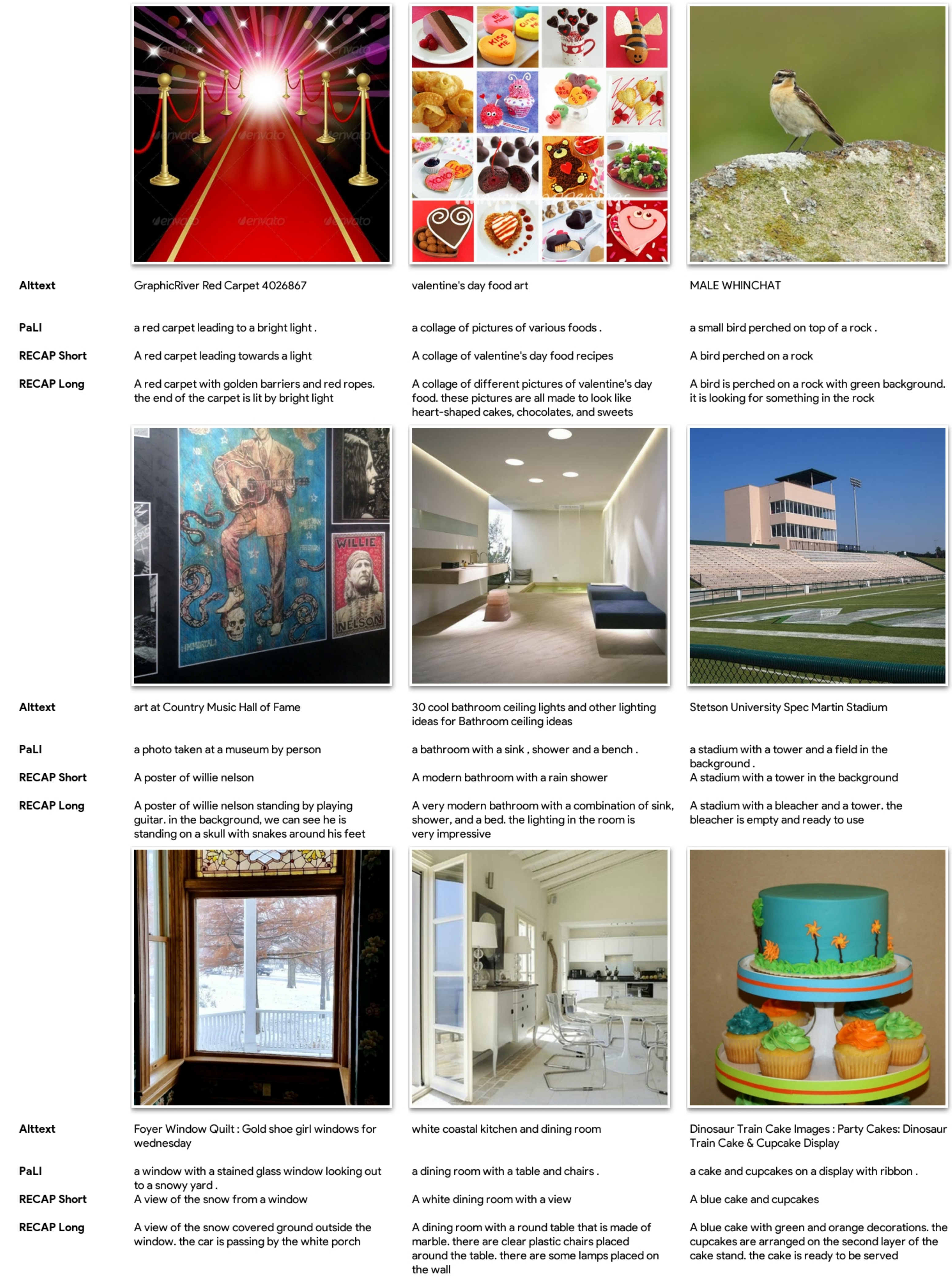}
\captionof{figure}{More examples of captions generated by RECAP variants vs. the original Alttext. Photos taken from LAION.}
    \label{fig:pali_outputs_table_2}
\end{figure*}

\twocolumn[{
\section{Additional Captioning Examples} \label{appendix_pali_outputs}

\cref{fig:pali_outputs_table_1} and \cref{fig:pali_outputs_table_2} show additional example captions generated by RECAP versus the original Alttext.

\begin{center}
    \includegraphics[width=0.88\linewidth]{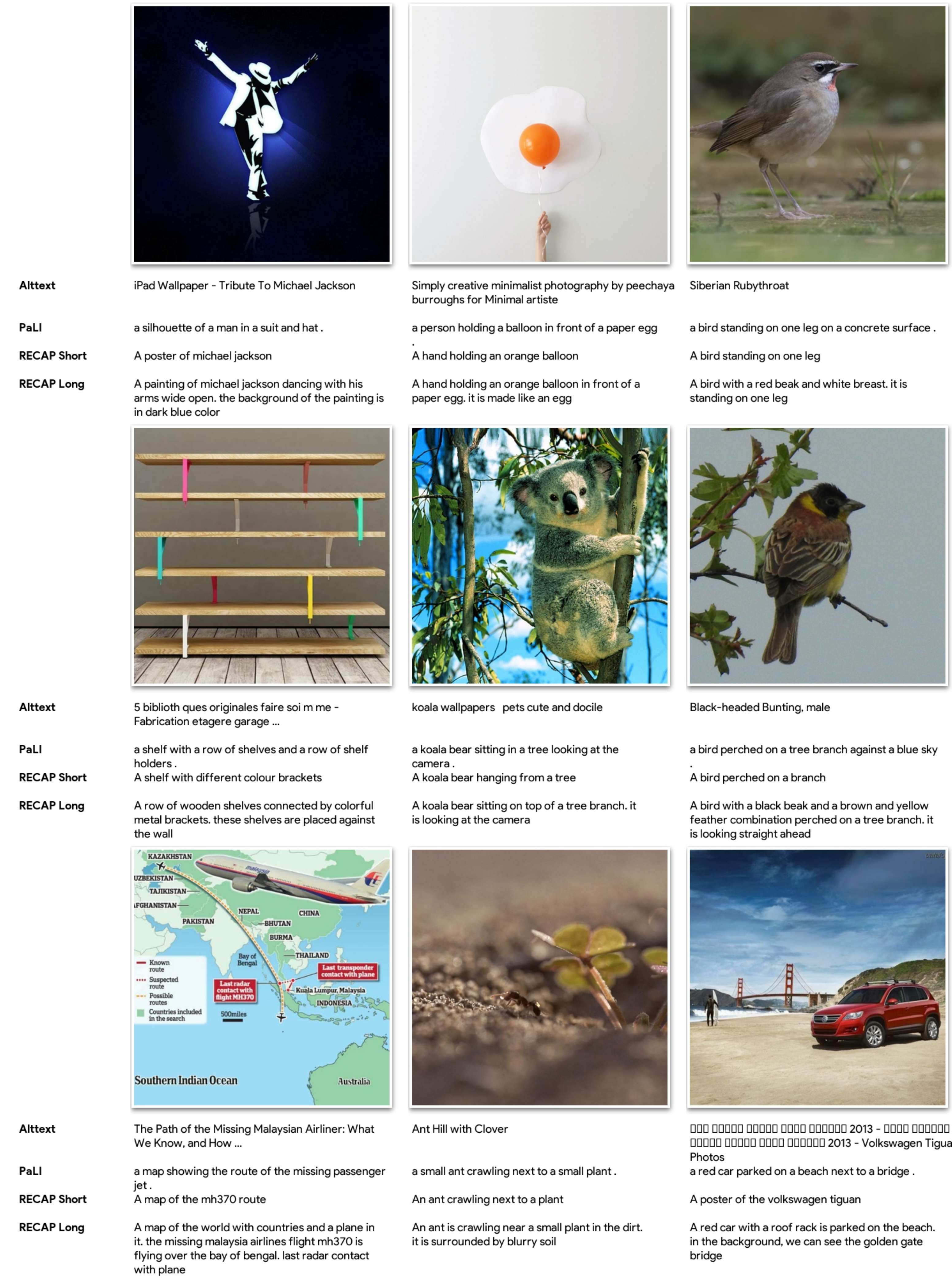}
\captionof{figure}{Examples of captions generated by RECAP variants vs. the original Alttext. Photos taken from LAION.}
    \label{fig:pali_outputs_table_1}
\end{center}
}]

\end{document}